\documentclass{article}
\usepackage{spconf,amsmath,graphicx}
\usepackage{bm}
\usepackage{enumitem}
\usepackage{amsmath}
\usepackage{algorithm}
\usepackage{algorithmic}
\usepackage{bbm}
\usepackage{xcolor}
\usepackage{booktabs}
 \usepackage{subfigure}
 \usepackage{url}
 
\title{Combining Self-Supervised and Supervised Learning with Noisy Labels}

\name{Yongqi Zhang$^{\dagger}$ \qquad Hui Zhang$^{\dagger}$ \qquad Quanming Yao$^{\star}$  \qquad Jun Wan$^{\natural}$}
\address{$^{\dagger}$ 4Paradigm Inc., Beijing, China\\ 
	$^{\star}$ Department of Electronic Engineering, Tsinghua University, Beijing, China \\ 
	$^{\natural}$ Institute of Automation, Chinese Academy of Sciences, Beijing, China}

\begin{document}
%
\maketitle
\begin{abstract}
Since convolutional neural networks (CNNs) can easily overfit noisy labels, 
which are ubiquitous in visual classification tasks, 
it has been a great challenge to train CNNs against them robustly. 
Various methods have been proposed for this challenge.
However,
none of them pay attention to the difference between representation and classifier learning of CNNs. 
Thus, 
inspired by the observation that classifier is more robust to noisy labels while representation is much more fragile,
and by the recent advances of self-supervised representation learning (SSRL) technologies, 
we design a new method,
i.e., CS$^3$NL, 
to obtain representation by SSRL without labels
and train the classifier directly with noisy labels.
Extensive experiments are performed on both synthetic and real benchmark datasets.
Results demonstrate that the proposed method
can beat the state-of-the-art ones by a large margin,
especially under a high noisy level.
\end{abstract}
\begin{keywords}
Convolutional neural network, noisy label learning, self-supervised learning, robustness
\end{keywords}
\section{Introduction}
\label{sec:intro}

Convolutional 
neural networks (CNNs)~\cite{lecun1998gradient,goodfellow2016deep}  
have achieved remarkable success in computer vision tasks such as
image classification~\cite{He2016DeepRL,szegedy2015going} and object detection~\cite{ren2015faster,he2017mask},
with large accurately annotated datasets like 
ImageNet~\cite{russakovsky2015imagenet} and COCO~\cite{lin2014microsoft}.
However, noisy labels are often ubiquitous and inevitable in real-world datasets.
Due to over-parameterization,
CNNs can easily overfit and memorize noisy labels,
leading to poor generalization~\cite{arpit2017closer,neyshabur2017exploring,zhang2016understanding,jiang2020beyond}. 
Thus,
how to robustly train CNNs against noisy labels is
an important problem.

Recently, a number of approaches have been proposed to 
robustly learn from noisy labels~\cite{han2020survey}.
There are approaches estimating how the labels can be corrupted~\cite{sukhbaatar2014training,xiao2015learning},
correcting and robustifying the loss \cite{patrini2017making,arazo2019unsupervised},
or filtering out noisy examples with memorization effect~\cite{arpit2017closer,jiang2018mentornet,han2018co}.
While both representation and classifier are jointly trained in CNNs,
existing works do not look inside what really happens to the two parts
when trained with noisy labels.

From classical statistics,
when classes are well-separated,
the classification can be accurate even under strong label noise.
This motivates us to decouple the training of representation and classifier.
Inspired by the fact that
classifiers are robust against noisy labels when a good representation is given,
and motivated by the recent advances in self-supervised learning (SSRL)~\cite{chen2020simple,he2020momentum},
which learn robust representations,
we propose to learn representations in a contrastive manner \cite{chaitanya2020contrastive}.
Specifically,
we introduce
CS$^3$NL that \underline{C}ombines \underline{S}elf-\underline{S}upervised and \underline{S}upervised learning with \underline{N}oisy \underline{L}abels.
The main contributions can be summarized as:
\begin{itemize}[leftmargin=10px, itemsep=1pt,topsep=3pt,parsep=0pt,partopsep=0pt]
	\item We observe that the representation learning and classifier learning have different behaviors, 
		thus decouple the training of two parts and not train the CNN end-to-end.
	\item We design a robust CNN training procedure that combines SSRL to improve 
		the robustness of representation learning.
	\item Experiments are performed on datasets 
	with both synthetic and real-world label noise,
	showing advantage over the SOTA by a large margin,
	especially at high noise levels.
\end{itemize}

\section{The proposed method}
\label{sec:method}

The existing CNN architectures can be split into two parts:
a backbone which encodes the input image into high-level representations,
and a classifier which projects the representation to predict the image labels.
We assume the classifier here has a simple structure like a linear projector.
From classical statistics,
the simpler classifiers are more robust to noisy labels
if the representations are well-separated~\cite{mitchell1997machine}.

For image classification,
when the labels are clean with little noise,
the representations learned in a supervised way can be better than unsupervised way.
But when the level of noisy labels gets higher,
the representations trained supervised  will become easily damaged,
while 
SSRL trained unsupervised
is able to learn good and robust representations.
In short,
simple classifier can exhibit robustness to noisy labels,
and SSRL can help learn better representations.

Based on these analysis,
we present a new training framework,
CS$^3$NL, which utilizes the robustness of the classifier to identify clean labels
and adaptively combines self-supervised learning and supervised learning with noisy labels.
Our framework,
illustrated in Figure~\ref{fig:overal_framework}, 
is based on a state-of-the-art SSRL model
BYOL~\cite{grill2020bootstrap},
which is introduced in Section~\ref{ssec:byol}.
And we propose to adapt BYOL with supervised learning with noisy label
in Section~\ref{ssec:designs}
and provide the training objective in Section~\ref{ssec:objective}.

\subsection{BYOL framework}
\label{ssec:byol}

\begin{figure}[!t] 
	\centering
	\includegraphics[width=1.0\columnwidth]{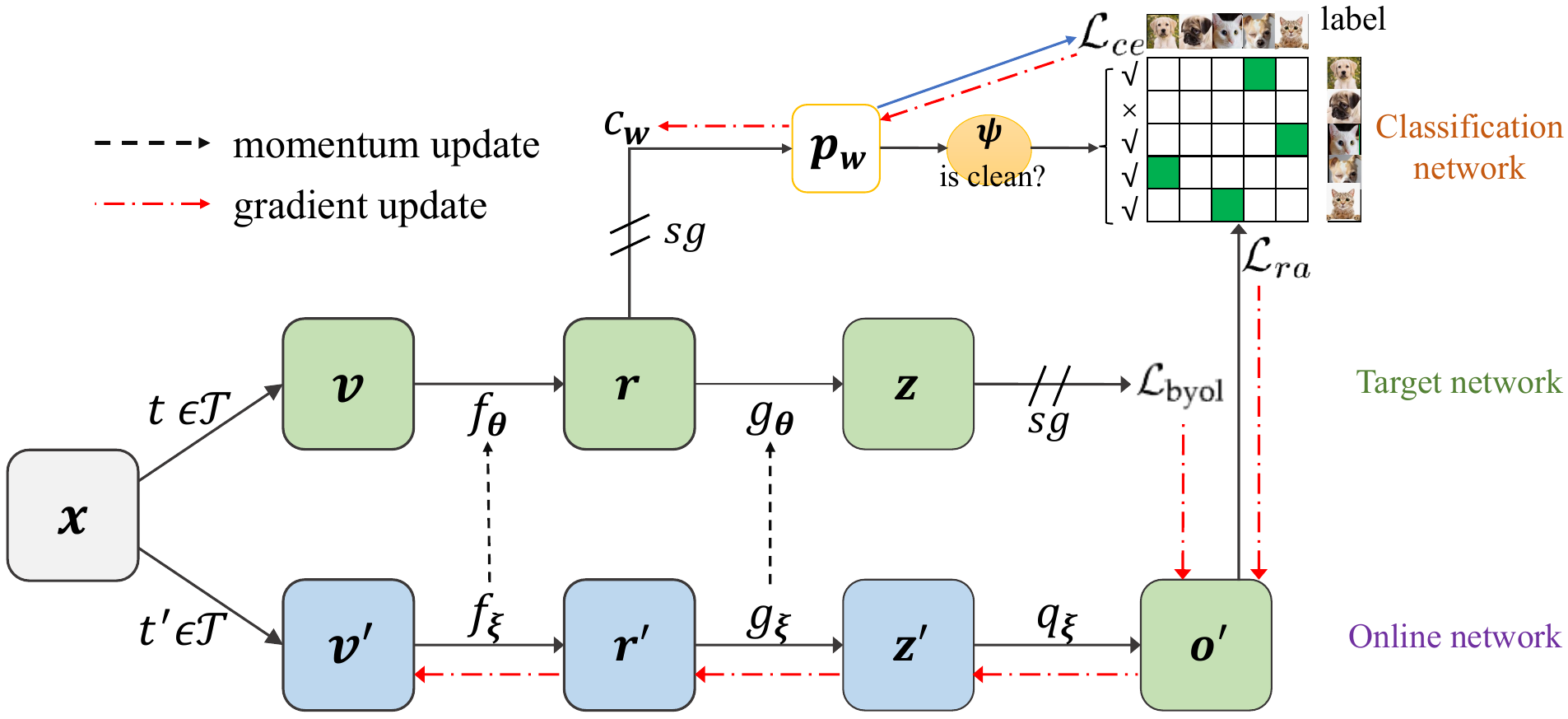}
	\vspace{-18px}
	\caption{Overall framework of our proposed method. Subscript $i$ is omitted here and
		\textit{sg} is short for stopping gradient.}
	\label{fig:overal_framework} 
	\vspace{-10px}
\end{figure}

Let $\bm{x}_i$ be an image,
$\bm{y}_i$ be its clean label, 
and $\bar{\bm{y}}_i$ be a noisy version of $\bm{y}_i$. 
A CNN usually
has two parts~\cite{lecun1998gradient,goodfellow2016deep}:
(i) Backbone $f$,
which stacks multiple convolutional and pooling layers,
and
extracts the representation for input image $\bm{x}_i$ as
$\bm{r}_i = f_{\bm \theta}(\bm{x}_i)$.
(ii) Classifier $c$,
which is usually a 
linear projector
with parameter $\bm{w}$
and produces predictions 
from the given representation as $c_{\bm w}(\bm{r}_i)$.
Thus, we denote a CNN as $C(\bm{x}_i) = c_{\bm w}(f_{\bm{\theta}}(\bm{x}_i))$.

BYOL is composed of the bottom two parts in Figure~\ref{fig:overal_framework},
i.e., an online network and a target network.
It first produces two augmented views 
$\bm{v}\!=\!t(\bm{x})$ and $\bm{v}^\prime\!=\!t^\prime(\bm{x})$ from $\bm{x}$ 
by applying twice random augmentation.
The view $\bm{v}'$ passes through an online network
with the backbone $f_{\bm{\xi}}$, 
the projector $g_{\bm{\xi}}$ 
and the predictor ${q}_{\bm{\xi}}$ in turn
to get features $\bm{r}'$, $\bm{z}'$ and $\bm{o}'$ respectively.
Another view $\bm{v}$ will be transformed to $\bm{r}$ and $\bm{z}$ by a similar target network,
which has no prediction module.
The target network's parameters $\bm{\theta}$ are an exponential moving average of the 
online parameters $\bm{\xi}$,
i.e., $\bm{\theta} = \tau \bm{\theta} + (1 - \tau) \bm{\xi}$ with momentum rate $\tau \in \left[0, 1\right]$.
The training loss of BYOL is defined as a mean squared error:
\begin{align}
	\label{eq:loss_byol}
	\mathcal{L}_{\text{byol}}(\bm{x}_i) 
	= \left\| \bm{o}'_{i} - \bm{z}_i \right\|^2_2.
\end{align}

\subsection{Dealing with noisy label}
\label{ssec:designs}

\subsubsection{Robust training of the classifier}
\label{ssec:robust_classifier}

To begin with, 
we add a linear classifier $c_{\bm{w}}$ 
on the top of the backbone $f_{\bm{\theta}}$ of target network 
(see Figure~\ref{fig:overal_framework}).
Different from the standard supervised learning schema,
we equip a $\mathsf{sg}$ (stop-gradient) operation 
between the backbone $f_{\bm{\theta}}$ and the classifier $c_{\bm{w}}$.
Thus,
we only use the gradients here
to update the classifier.
The objective for the classification branch is
\begin{equation}
	\mathcal{L}_{\text{ce}}
	(\bm{x}_i)  = 
	\mathsf{CE} 
	\Big( 
	c_{\bm{w}} 
	\big[
	\mathsf{sg} \left(f_{\bm{\theta}}(t(\bm{x}_i))\right) 
	\big], 
	\bar{\bm{y}}_i 
	\Big),
	\label{eq:loss_ce}
\end{equation}
where $\mathsf{CE}(\cdot, \cdot)$ denotes the cross-entropy loss function and 
the training sample $(\bm{x}_i, \bar{\bm{y}}_i)$ is drawn from a noisy dataset.

The $\mathsf{sg}$ operator is very important here.
We know that 
the classifier is more robust than the representation 
in presence of noisy labels,
and
the representation is inherent from 
the target network,
which can capture some distinct patterns in images by SSRL.
Thus,
the gradients from $\mathcal{L}_{\text{ce}}$ can help train the classifier
to predict correct labels,
and the $\mathsf{sg}$ operator
avoids the gradients from noisy labels affecting the representation.
These will also been
empirically shown in Section~\ref{exp:effectiveness_of_cs3nl}.

\subsubsection{Identify clean labels}

Next,
since the classifier can predict correct labels,
we can roughly separate the training samples into 
two groups,
i.e.,
a group whose labels
are likely to be correct (denote as $\mathsf{L}$)
and
another group whose labels
are likely to be wrong
(denote as $\mathsf{U}$).
Inspired by~\cite{han2018co,li2020dividemix},
we cluster samples based on their loss,
as small loss samples are more likely to have clean labels. 
We use a two-components Gaussian mixture model (GMM) \cite{reynolds2009gaussian}  $\psi$
to measure whether the sample is more likely to have clean label.
In this way,
the classification branch helps us
find clean labels from the training samples.

\subsubsection{Combine clean labels with SSRL}
\label{ssec:rep_alignment}

We know that
once the labels are clean enough,
we can use the clean labels to improve the representation learned by SSRL.
Here, 
we introduce a representation alignment loss
$\mathcal{L}_{\text{ra}}$, 
to assist the SSRL.
Specifically, 
$\mathcal{L}_{\text{ra}}$ is defined as
\begin{align} 
	\mathcal{L}_{\text{ra}}
	(\bm{x}_i) 
	\! = \! \sum\nolimits_{j=1}^{N_L}
	\mathbbm{1}_{j \neq i} 
	\mathbbm{1}_{\bm{x}_i, \bm{x}_j \in \mathsf{L}} 
	\mathbbm{1}_{\bar{\bm{y}}_i = \bar{\bm{y}}_j}
	\left\| \bm{o}'_i - \bm{z}_i \right\|^2_2,
	\label{eq:loss_ra}
\end{align}
where 
$\mathbbm{1}_{\text{condition}} \!\in\! \{0,1\}$ is an indicator function 
evaluating $1$ iff condition is true,
$N_L$ is the number of samples in $\mathsf{L}$.
Different from~\eqref{eq:loss_byol} that 
treats two different augmented views
from the same image $\bm{x}_i$ as a positive pair ($\bm{v}_i = t(\bm{x}_i), \bm{v}_i^\prime = t^\prime(\bm{x}_i)$).
\eqref{eq:loss_ra} considers more positive pairs from different images but sharing the same label,
and make sure that only
clean labeled data (i.e., $\mathsf{L}$) can be used here to 
avoid bad effect of noise on representation learning.

\subsection{Training details}
\label{ssec:objective}
Many state-of-the-art methods~\cite{li2020dividemix,han2018co,zhang2017mixup,han2019deep}
get excellent results with the help of 
image Mixup and model ensembling.
Image Mixup is a simple learning technique 
proposed in~\cite{zhang2017mixup}, 
which trains a neural network on the combination of 
pairs of images and their labels.
Model ensembling ensembles the model output of two separate neural networks
to improve generalization.
Due to their superiority and simple usage,
we also use the two techniques to enhance CS$^3$NL here.

Finally, the total training loss can be formulated as 
the combination of losses in \eqref{eq:loss_byol}, \eqref{eq:loss_ce} and \eqref{eq:loss_ra},
i.e.,
\begin{equation}
	\mathcal{L} = 
	\sum\nolimits_{i=1}^N
	\mathcal{L}_{\text{byol}}(\bm{x}_i) + 
	\lambda_1 \mathcal{L}_{\text{ra}}(\bm{x}_i) + 
	\lambda_2 \mathcal{L}_{\text{ce}}(\bm{x}_i),
	\label{eq:training_obj}
\end{equation}
on the $N$ images $\bm x_i$ in the training set.
$\lambda_1$ and $\lambda_2$
weight the importance between these three losses.
In our experiments, 
we set $\lambda_1$ and $\lambda_2$ to 1.
The model parameters are optimized by stochastic gradient descent
with mini-batches of datasets.


\section{Experiments}
In this section, 
experiments are performed on three popular benchmark datasets
\cite{li2020dividemix}:
CIFAR-10, CIFAR-100~\cite{krizhevsky2009learning}
and WebVison~\cite{li2017webvision}.
All algorithms are implemented in PyTorch, and run on a linux server with 8 NVIDIA RTX 3090 GPUs.

\subsection{Synthetic noisy labels}
\label{sec:data:cifar}

Following~\cite{han2018co,li2020dividemix}, 
we add both \textit{symmetric} and \textit{asymmetric} noise 
on CIFAR-10, and only \textit{symmetric} noise on CIFAR-100
with different levels of noises.
The proposed CS$^3$NL is compared 
with the following methods:
\begin{itemize}[leftmargin=*,itemsep=1pt,topsep=2pt,parsep=0pt,partopsep=1pt]
	\item  Standard: directly trains the network with noisy labels;
	\item Mixup~\cite{zhang2017mixup}:
	the image {mixup} technique on top of Standard.
	\item F-correction~\cite{patrini2017making}, Meta-Learn~\cite{li2019learning} and M-correction~\cite{arazo2019unsupervised}:
	correct the training objects or select the clean samples to robustify training;
	\item Co-teaching~\cite{han2018co}:
	trains two networks and each network selects clean samples for training the other,
	and its extensions DivideMix~\cite{li2020dividemix}, ELR+~\cite{liu2020early}
	AugDesc~\cite{nishi2021augmentation}.
\end{itemize}
In particular,
M-correction, DivideMix, ELR+, AugDesc are enhanced with the Mixup technique.
Currently,
ELR+ and AugDesc are the state-of-the-arts.
Following ~\cite{li2020dividemix,liu2020early},
PreAct-ResNet18 is used as backbone for all the methods.


Table~\ref{tab:cifar_sym} shows the highest testing accuracies across
training
epochs.
We observe that methods utilizing memorization effect in the third part
are better then methods correcting the noisy labels in second parts.
The \textit{mixup} technique is beneficial 
when applied to different methods.
DivideMix and ELR+
have comparable performance,
showing
the benefits of combining
clean sample selection by two joint networks.
AugDesc is better than other baselines.
The proposed 
CS$^3$NL outperforms
all competing methods. 
The improvement is particularly significant on
higher noise. 
For example,
at the noise level of 90\%, 
the relative improvement over the best baseline is
7.2\% 
on CIFAR-10, 
and 18.6\% on CIFAR-100.
This demonstrate the significance of SSRL.


\begin{table}[!t]
	\centering
	\caption{Testing accuracies  (\%) on the CIFAR-10 and CIFAR-100 datasets.
		Results of the baselines are from~\cite{li2020dividemix,liu2020early},
		where``-'' means the corresponding result is not available.
		The highest accuracy is in bold.}
	\setlength\tabcolsep{1.3pt}
	\small
	\begin{tabular}{c |c|c|c|c|c|c|c|c|c}
		\toprule
		&   \multicolumn{5}{c|}{CIFAR-10}                  &  \multicolumn{4}{c}{CIFAR-100}      \\ 
		&  \multicolumn{4}{c|}{sym noise}  & asym			  &  \multicolumn{4}{c}{sym noise}			\\ 
		& 20\% & 50\% & 80\% & 90\%  & 40\% &       20\%       &       50\%       &     80\%      &     90\%      \\ \midrule
		Standard                                    & 86.8 & 79.4 & 62.9 & 42.7  & 85.0 &       62.0       &       46.7       &     19.9      &     10.1      \\ 
		Mixup~\cite{zhang2017mixup}        & 95.6 & 87.1 & 71.6 & 52.2  & -    &       67.8       &       57.3       &     30.8      &     14.6      \\
		\midrule
		F-correction~\cite{patrini2017making}       & 86.8 & 79.8 & 63.3 & 42.9  & 87.2 &       61.5       &       46.6       &     19.9      &     10.2      \\
		Meta-Learn~\cite{li2019learning}                   & 92.9 & 89.3 & 77.4 & 58.7  & 89.2 &       68.5       &       59.2       &     42.4      &     19.5      \\ 
		M-correction~\cite{arazo2019unsupervised}   & 94.0 & 92.0 & 86.8 & 69.1  & 87.4 &       73.9       &       66.1       &     48.2      &     24.3      \\  \midrule
		Co-teaching~\cite{han2018co}            & 89.5 & 85.7 & 67.4 & 47.9  & -    &       65.6       &       51.8       &     27.9      &     13.7      \\
		DivideMix~\cite{li2020dividemix}            & 96.1 & 94.6 & 93.2 & 76.0  & 93.4 &       77.3       &       74.6       &     60.2      &     31.5      \\
		ELR+~\cite{liu2020early}					& 95.8 & 94.8 & 93.3 & 78.7  & 93.0 &       77.6       &       73.6       &     60.8      &     33.4      \\ 
		AugDesc~\cite{nishi2021augmentation}			& \textbf{96.3} & 95.1 & 93.8 & 83.9  & 94.4 &       \textbf{79.5}       &       {75.2}       &     64.4      &     41.2      \\  \midrule
		CS$^3$NL	 & \textbf{96.3} & \textbf{95.2} & \textbf{94.4}& \textbf{89.9}  & \textbf{94.7} &       {78.2}       &       \textbf{76.7}       &     \textbf{67.0}      &     \textbf{48.9}      \\
		\hline
	\end{tabular}
	\label{tab:cifar_sym}
	\vspace{-10px}
\end{table}

\subsection{Real-world noisy labels}

Following~\cite{jiang2018mentornet,chen2019understanding}, 
we only compare on the mini WebVision dataset \cite{li2017webvision},
which contains the top 50 classes from the ImageNet ILSVRC12 subsets~\cite{russakovsky2015imagenet},
and has about 66K noisy samples for training and 50K clean samples for validation.

As in~\cite{li2020dividemix,liu2020early},
we use the InceptionResNetV2 network as backbone
and compare CS$^3$NL with the following methods:
F-correction~\cite{patrini2017making},
Co-teaching~\cite{han2018co},
DivideMix~\cite{li2020dividemix},
and ELR+~\cite{liu2020early}.
Besides, we also compare with
Iterative-CV~\cite{chen2019understanding}, which identifies clean samples by
cross-validation;
MentorNet~\cite{jiang2018mentornet}, which trains a student network on samples weighed by a pre-trained teacher network;
D2L~\cite{ma2018dimensionality}, which linearly combines label and model
predictions.
Following~\cite{li2020dividemix,liu2020early}, 
the models are evaluated 
with Top-1 and Top-5 prediction accuracies in 
Table~\ref{tab:webvision_results}.
Again,
DivideMix and ELR+ are very competitive in the real-word noisy labels,
but the proposed method outperforms all consistently on both datasets.

\begin{table}[!t]
	\caption{Comparison with baseline methods on WebVision and ILSVRC12. Results of others are copied from~\cite{li2020dividemix,liu2020early}.}
	\setlength\tabcolsep{3.3pt}
	\centering
	\small
	\begin{tabular}{c|c|c|c|c}
		\toprule
		& \multicolumn{2}{c|}{WebVision} & \multicolumn{2}{c}{ILSVRC12} \\
		& Top-1 (\%)         & Top-5 (\%)         & Top-1 (\%)         & Top-5 (\%)       		\\ \midrule
		F-correction~\cite{patrini2017making}    & 61.12          & 82.68          & 57.36          & 82.36        		\\
		Co-teaching~\cite{han2018co}             & 63.58          & 85.20          & 61.48          & 84.70        		\\
		DivideMix~\cite{li2020dividemix}         & 77.32          & 91.64          & {75.20}          & {90.84}        		\\
		ELR+~\cite{liu2020early}			     & {77.78}          & {91.68}          & 70.29          & 89.76        		\\
		Iterative-CV~\cite{chen2019understanding}& 65.24          & 85.34          & 61.60          & 84.98        		\\
		MentorNet~\cite{jiang2018mentornet}      & 63.00          & 81.40          & 57.80          & 79.92        		\\
		D2L~\cite{ma2018dimensionality}          & 62.68          & 84.00          & 57.80          & 81.36        		\\
		\midrule
		CS$^3$NL                                     & \textbf{79.12} & \textbf{93.00} & \textbf{77.20} & \textbf{93.04}    \\ 
		\bottomrule
	\end{tabular}
	\label{tab:webvision_results}
	\vspace{-8px}
\end{table}

\subsection{Ablation studies}
\label{exp:effectiveness_of_cs3nl}

To evaluate the effectiveness 
of our novel refinements proposed in Section~\ref{sec:method},
we perform a serial of ablation studies 
for different training strategies (i.e., (i)-(v)) 
on the CIFAR10 dataset with ResNet34 as the backbone. 
Meanwhile, 
we keep the same hyperparameters setting 
as described in Section~\ref{sec:data:cifar} 
for a fair comparison. 
Medium (40\%) and high (80\%) levels of symmetric noise 
are used, following~\cite{liu2020early}. 
Besides, 
to verify the upper bound performance 
of our proposed jointly training approach,
we further perform experiments with a clean CIFAR10 dataset (i.e., 0\%). 
To fully understand 
the significance of SSRL,
a purely supervised learning baseline (vi))
is also included here. 
Six models are compared:
\begin{enumerate}[leftmargin=18pt,itemsep=1pt,topsep=3pt,parsep=0pt,partopsep=1pt]
	\item[(i)] 
	CS$^3$NL: The proposed method.
	
	\item[(ii)]  
	Online CS$^3$NL: Based on (i),
	instead of training classifier $c_{\bm w}$ 
	on the top of backbone module from the target network, 
	we choose the backbone module from the online network;
	
	\item[(iii)]  
	w/o $\mathsf{sg}$ in \eqref{eq:loss_ce}: Based on (i),
	we discard the stop-gradient operation in \eqref{eq:loss_ce};
	
	\item[(iv)] 
	w/o $\psi$: Based on (i),
	we discard the label cleaning module $\psi$ and 
	regard all training data as $\mathsf{L}$;
	
	\item[(v)] 
	w/o \eqref{eq:loss_ra}: Based on (i),
	we exclude the SSRL loss in \eqref{eq:loss_ra} 
	from the total loss in \eqref{eq:training_obj};
	
	\item[(vi)]
	Standard: we train a CNN 
	in an end-to-end learning manner 
	using the standard cross-entropy loss.
\end{enumerate}

From Table~\ref{tab:effectiveness},
we observe that
changing target network to online network has little influence as their parameters are shared.
Stop gradient is the most important part here 
since it decouples the representation learning and classifier learning.
In addition,
all of the design components can improve the robustness against noisy label
compared with standard setting.

\begin{table}[!t]
	\caption{Validation experiments for evaluating the effectiveness of our novel refinements of CS$^3$NL. 
	The values in parentheses mean the absolute gap to CS$^3$NL}
	\centering
	\small
	\setlength\tabcolsep{5pt}
	\begin{tabular}{c|l|c|c|c}
		\toprule
		& Setting                                  & 0\%                     & 40\%                    & 80\%                   \\ \midrule 
		(i)  & CS$^3$NL                                 & 96.9          & 94.6        & 88.1      \\ \midrule 
		(ii) & Online CS$^3$NL                          & 95.6 (-1.3)      & 94.1 (-0.5)       & 87.6 (-0.5)    \\  
		(iii)& w/o $\mathsf{sg}$ in \eqref{eq:loss_ce}  & 95.8 (-1.1)       & 91.4 (-3.2)       & 80.7 (-7.4)     \\  
		(iv) & w/o $\psi$                               & 95.7 (-1.2)        & 92.5 (-2.1)       & 86.2 (-1.9)      \\ 
		(v)  & w/o \eqref{eq:loss_ra}                   & 94.3 (-2.6)        & 93.6 (-1.0)    & 87.9 (-0.2)      \\ \midrule
		(vi) & Standard		                                & 96.8 (-0.1)       & 81.3 (-13.3)   & 62.9 (-25.2)     \\ \bottomrule 
	\end{tabular}
	\label{tab:effectiveness}
	\vspace{-5px}
\end{table}

\section{RELATION TO PRIOR WORKS}
\label{sec:related}

\subsection{Noisy label learning}
The methods studying the robustness of training CNNs with noisy labels
can be categorized into three directions.
The first type estimates a noisy transition matrix on the noisy labels~\cite{sukhbaatar2014training,xiao2015learning},
but the transition matrix is hard to estimate correctly.
The second type modifies or redesigns the training loss
to improve the robustness \cite{patrini2017making,arazo2019unsupervised}.
However, as the CNNs are often over-parameterized,
the influence of noisy label still exists given sufficient training time \cite{zhang2016understanding}.
The third type conducts label division based on the memorization effect,
which tries to distinguishes the noise samples, discard or reuse them~\cite{arpit2017closer,jiang2018mentornet,han2018co}.
Different from these approaches,
we explore CNN training mechanism with noisy labels 
from two perspectives,
i.e.,
classifier learning
and representation learning.
And we combine SSRL and 
supervised learning with noisy labels
to make both parts better.

\subsection{Semi-supervised representation learning (SSRL)}
\label{ssec:selfrep}
SSRL~\cite{chen2020simple,he2020momentum}
is a kind of unsupervised learning method where no labels are required.
Recently,
SSRL achieves state-of-the-art performance in unsupervised training of neural networks
by contrastive proxy task \cite{chaitanya2020contrastive},
which maximizes (\textit{resp}. minimizes) the intra- (\textit{resp}. inter-) class distances.
In this work, we adopt the BYOL framework \cite{grill2020bootstrap}
which achieves remarkable performance.
While SSRL can help learn better representations,
it does not utilize any label information which can contain more meaningful visual semantics.
By observing the learning behavior of representation learning and classifier learning
as well as that of SSRL and supervised learning with noisy labels,
we are the first to 
combine SSRL and supervised learning to enhance the CNN's robustness to noisy labels.
The new framework in Figure~\ref{fig:overal_framework}
has significant advance in the noisy label learning problem.

\section{Conclusion}

In this paper, 
we look deeply into the robustness of representation and classifier learning in the presence of noisy labels.
We find that noisy labels will damage the representation learning significantly than classifier learning, 
and the classifier itself can exhibit strong robustness w.r.t. noisy labels with a good representation.  
Motivated by this, 
we proposed a robust CNN 
training manner to take care of both representation learning and classifier learning.
By combining it with the current SOTA self-supervised representation learning framework, 
a large-scale improvement has been achieved.




\bibliographystyle{IEEEbib}
\bibliography{refs}

\begin{thebibliography}{10}

\bibitem{lecun1998gradient}
Y.~LeCun, L.~Bottou, Y.~Bengio, and P.~Haffner,
\newblock ``Gradient-based learning applied to document recognition,''
\newblock {\em Proceedings of the IEEE}, vol. 86, no. 11, pp. 2278--2324, 1998.

\bibitem{goodfellow2016deep}
I.~Goodfellow, Y.~Bengio, and A.~Courville,
\newblock {\em Deep learning},
\newblock MIT press, 2016.

\bibitem{He2016DeepRL}
K.~He, X.~Zhang, S.~Ren, and J.~Sun,
\newblock ``Deep residual learning for image recognition,''
\newblock in {\em CVPR}, 2016.

\bibitem{szegedy2015going}
C.~Szegedy, W.~Liu, Y.~Jia, et~al.,
\newblock ``Going deeper with convolutions,''
\newblock in {\em CVPR}, 2015, pp. 1--9.

\bibitem{ren2015faster}
S.~Ren, K.~He, R.~Girshick, and J.~Sun,
\newblock ``Faster r-cnn: Towards real-time object detection with region
  proposal networks,''
\newblock {\em NeurIPS}, vol. 28, 2015.

\bibitem{he2017mask}
K.~He, G.~Gkioxari, P.~Doll{\'a}r, and R.~Girshick,
\newblock ``Mask {R-CNN},''
\newblock in {\em ICCV}, 2017.

\bibitem{russakovsky2015imagenet}
O.~Russakovsky, J.~Deng, H.~Su, et~al.,
\newblock ``Imagenet large scale visual recognition challenge,''
\newblock {\em IJCV}, vol. 115, pp. 211--252, 2015.

\bibitem{lin2014microsoft}
T.~Lin, M.~Maire, S.~Belongie, et~al.,
\newblock ``Microsoft coco: Common objects in context,''
\newblock in {\em ECCV}. Springer, 2014, pp. 740--755.

\bibitem{arpit2017closer}
D.~Arpit, S.~Jastrzbski, N.~Ballas, et~al.,
\newblock ``A closer look at memorization in deep networks,''
\newblock in {\em ICML}. PMLR, 2017, pp. 233--242.

\bibitem{neyshabur2017exploring}
B.~Neyshabur, S.~Bhojanapalli, D.~McAllester, and N.~Srebro,
\newblock ``Exploring generalization in deep learning,''
\newblock in {\em NeurIPS}, 2017.

\bibitem{zhang2016understanding}
C.~Zhang, S.~Bengio, M.~Hardt, B.~Recht, and O.~Vinyals,
\newblock ``Understanding deep learning requires rethinking generalization,''
\newblock in {\em ICLR}, 2017.

\bibitem{jiang2020beyond}
L.~Jiang, D.~Huang, M.~Liu, and W.~Yang,
\newblock ``Beyond synthetic noise: Deep learning on controlled noisy labels,''
\newblock in {\em ICML}. PMLR, 2020, pp. 4804--4815.

\bibitem{han2020survey}
B.~Han, Q.~Yao, T.~Liu, G.~Niu, I.~W Tsang, J.~T Kwok, and M.~Sugiyama,
\newblock ``A survey of label-noise representation learning: Past, present and
  future,''
\newblock Tech. {R}ep., arXiv:2011.04406, 2020.

\bibitem{sukhbaatar2014training}
S.~Sukhbaatar, J.~Bruna, M.~Paluri, L.~Bourdev, and R.~Fergus,
\newblock ``Training convolutional networks with noisy labels,''
\newblock in {\em ICLR}, 2015.

\bibitem{xiao2015learning}
T.~Xiao, T.~Xia, Y.~Yang, C.~Huang, and X.~Wang,
\newblock ``Learning from massive noisy labeled data for image
  classification,''
\newblock in {\em CVPR}, 2015.

\bibitem{patrini2017making}
G.~Patrini, A.~Rozza, A.~Krishna~Menon, R.~Nock, and L.~Qu,
\newblock ``Making deep neural networks robust to label noise: A loss
  correction approach,''
\newblock in {\em CVPR}, 2017.

\bibitem{arazo2019unsupervised}
E.~Arazo, D.~Ortego, P.~Albert, N.~O'Connor, and K.~McGuinness,
\newblock ``Unsupervised label noise modeling and loss correction,''
\newblock in {\em ICML}, 2019.

\bibitem{jiang2018mentornet}
J.~Lu, Z.~Zhou, T.~Leung, et~al.,
\newblock ``Mentornet: Learning data-driven curriculum for very deep neural
  networks on corrupted labels,''
\newblock in {\em ICML}. PMLR, 2018, pp. 2304--2313.

\bibitem{han2018co}
B.~Han, Q.~Yao, X.~Yu, G.~Niu, M.~Xu, W.~Hu, I.~Tsang, and M.~Sugiyama,
\newblock ``Co-teaching: Robust training of deep neural networks with extremely
  noisy labels,''
\newblock in {\em NeurIPS}, 2018.

\bibitem{chen2020simple}
T.~Chen, S.~Kornblith, M.~Norouzi, and G.~Hinton,
\newblock ``A simple framework for contrastive learning of visual
  representations,''
\newblock in {\em ICML}, 2020.

\bibitem{he2020momentum}
K.~He, H.~Fan, Y.~Wu, S.~Xie, and R.~Girshick,
\newblock ``Momentum contrast for unsupervised visual representation
  learning,''
\newblock in {\em CVPR}, 2020.

\bibitem{chaitanya2020contrastive}
K.~Chaitanya, E.~Erdil, N.~Karani, and E.~Konukoglu,
\newblock ``Contrastive learning of global and local features for medical image
  segmentation with limited annotations,''
\newblock {\em NeurIPS}, vol. 33, pp. 12546--12558, 2020.

\bibitem{mitchell1997machine}
T.~M Mitchell,
\newblock {\em Machine learning}, vol.~1,
\newblock McGraw-hill New York, 1997.

\bibitem{grill2020bootstrap}
J.~Grill, F.~Strub, F.~Altch{\'e}, et~al.,
\newblock ``Bootstrap your own latent-a new approach to self-supervised
  learning,''
\newblock {\em NeurIPS}, vol. 33, pp. 21271--21284, 2020.

\bibitem{li2020dividemix}
J.~Li, R.~Socher, and S.~Hoi,
\newblock ``Dividemix: Learning with noisy labels as semi-supervised
  learning,''
\newblock in {\em ICLR}, 2020.

\bibitem{reynolds2009gaussian}
D.~A Reynolds et~al.,
\newblock ``Gaussian mixture models.,''
\newblock {\em Encyclopedia of biometrics}, vol. 741, no. 659-663, 2009.

\bibitem{zhang2017mixup}
H.~Zhang, M.~Cisse, Y.~Dauphin, and D.~Lopez-Paz,
\newblock ``{MixUp}: Beyond empirical risk minimization,''
\newblock in {\em ICLR}, 2018.

\bibitem{han2019deep}
J.~Han, P.~Luo, and X.~Wang,
\newblock ``Deep self-learning from noisy labels,''
\newblock in {\em ICCV}, 2019.

\bibitem{krizhevsky2009learning}
A.~Krizhevsky, G.~Hinton, et~al.,
\newblock ``Learning multiple layers of features from tiny images,''
\newblock 2009.

\bibitem{li2017webvision}
W.~Li, L.~Wang, W.~Li, E.~Agustsson, and L.~Van~Gool,
\newblock ``Webvision database: Visual learning and understanding from web
  data,''
\newblock Tech. {R}ep., 2017.

\bibitem{li2019learning}
J.~Li, Y.~Wong, Q.~Zhao, and M.~S Kankanhalli,
\newblock ``Learning to learn from noisy labeled data,''
\newblock in {\em CVPR}, 2019, pp. 5051--5059.

\bibitem{liu2020early}
S.~Liu, J.~Niles-Weed, N.~Razavian, and C.~Fernandez-Granda,
\newblock ``Early-learning regularization prevents memorization of noisy
  labels,''
\newblock {\em NeurIPS}, vol. 33, pp. 20331--20342, 2020.

\bibitem{nishi2021augmentation}
K.~Nishi, Y.~Ding, A.~Rich, and T.~Hollerer,
\newblock ``Augmentation strategies for learning with noisy labels,''
\newblock in {\em CVPR}, 2021, pp. 8022--8031.

\bibitem{chen2019understanding}
P.~Chen, B.~Liao, G.~Chen, and S.~Zhang,
\newblock ``Understanding and utilizing deep neural networks trained with noisy
  labels,''
\newblock in {\em ICML}, 2019.

\bibitem{ma2018dimensionality}
X.~Ma, Y.~Wang, M.~E Houle, et~al.,
\newblock ``Dimensionality-driven learning with noisy labels,''
\newblock in {\em ICML}, 2018.

\end{thebibliography}

\end{document}